\documentclass[11pt]{article}
\usepackage{acl}
\usepackage{times}
\usepackage{latexsym}
\usepackage[T1]{fontenc}
\usepackage[utf8]{inputenc}
\usepackage{microtype}
\usepackage{graphicx}
\usepackage{booktabs}
\usepackage{tabularx}
\usepackage{threeparttable}
\newcolumntype{L}{>{\raggedright\arraybackslash}X}
\newcolumntype{C}{>{\centering\arraybackslash}X}
\usepackage{amsmath,amssymb}
\usepackage{tcolorbox}
\tcbuselibrary{skins}

\definecolor{mygreen}{rgb}{0.27,0.70,0.23}   
\definecolor{myyellow}{rgb}{0.74,0.71,0.02}  
\definecolor{cadmiumgreen}{rgb}{0.0, 0.42, 0.24}
\definecolor{myred}{rgb}{0.7, 0.3, 0.0}
\definecolor{myblue}{rgb}{0.2, 0.3, 0.6}

\title{FlyRoute: Self-Evolving Agent Profiling via Data Flywheel for Adaptive Task Routing}

\author{Rongjun Li \quad Ziyu Zhou \quad Yihang Wu \\
  IT Innovation and Research Center, Huawei Technologies \\
  \texttt{\{lirongjun3, zhouziyu8\}@huawei.com} \quad
  \texttt{wuyihang2@h-partners.com}}

\begin{document}
\maketitle
\begin{abstract}
Enterprise routers assign queries to expert agents, yet deployed agent profiles often become stale as prompts, tools and underlying models evolve, while developers rarely maintain descriptions or exemplars after registration. We present FlyRoute, a continual agent-profiling framework that grows capability evidence from deployment traffic: the router dispatches candidate agents, a quality-gate retains only satisfactory query-agent interactions as profiling evidence, accumulated evidence is periodically distilled into updated capability descriptions, and those descriptions are combined with BM25-retrieved successes in an LLM router. To make this flywheel data-efficient, FlyRoute introduces a targeted exploration policy that combines profile-coverage uncertainty, BM25 relevance, and lexical novelty, prioritizing under-profiled agents only for plausible queries while reducing redundant evidence collection. Importantly, profile construction relies solely on routed interactions and quality-gated responses. In experiments on a proprietary enterprise developer-support benchmark of real routed queries, FlyRoute improves a same-backbone zero-shot LLM router from 72.57\% to 78.04\% with only five seed queries per agent, showing that profile retrieval already strengthens cold-start routing. After replaying 7{,}211 training queries through the flywheel, accuracy rises to 89.83\% (+17.26pp over zero-shot; +11.79pp over cold start), with consistent gains across four expert domains under standard single-gold routing accuracy. These results suggest that continually updated profiles can substantially improve routing quality without retraining the underlying router.
\end{abstract}

\section{Introduction}
\label{sec:intro}

Multi-Agent Systems (MAS) have emerged as a powerful paradigm for building intelligent applications by decomposing complex problems into sub-tasks handled by specialized agents~\citep{park2023generative}. Within these systems, the router---responsible for determining which expert agent should handle a given query---plays a crucial role in overall performance. Existing routing methods, whether based on supervised encoder routers~\citep{simonds2024modem}, embedding similarity~\citep{piskala2025dynamic}, or structure-based learning methods~\citep{zhao2026tcar}, share a common assumption: that each agent's capabilities can be accurately described at registration time and will remain stable thereafter.

This assumption breaks down in real-world enterprise deployments for two reasons. First, agent developers often cannot provide accurate capability descriptions. In practice, an agent's behavior is determined by its system prompt, the tools it has access to, and the underlying model---all of which are under active development. Writing a precise natural-language description of what an agent can and cannot do is time-consuming, and even when provided, such descriptions quickly become stale as the agent evolves. Representative examples are equally hard to curate: developers may not know which queries best characterize the agent's competence boundaries. Second, agent capabilities change after deployment. A developer may update the agent's prompt to handle new scenarios, integrate a new tool that extends its reach, or switch to a more capable model. When this happens, the original profile registered with the router no longer reflects the agent's actual behavior, yet the router continues to make decisions based on outdated information.

We observe that the root cause of both problems is the static nature of current agent profiling: capabilities are typically captured at registration and seldom revised as agents evolve in production. This motivates a different approach: continuously updating agent profiles from successful deployment interactions. When a query is routed to an agent and the resulting response passes a quality gate, that interaction provides empirical evidence about the agent's capabilities. Over time, accumulated evidence forms a richer and more current profile than manually maintained descriptions because it reflects observed behavior rather than intended behavior.

\begin{figure}[t]
  \centering
  \includegraphics[width=\linewidth]{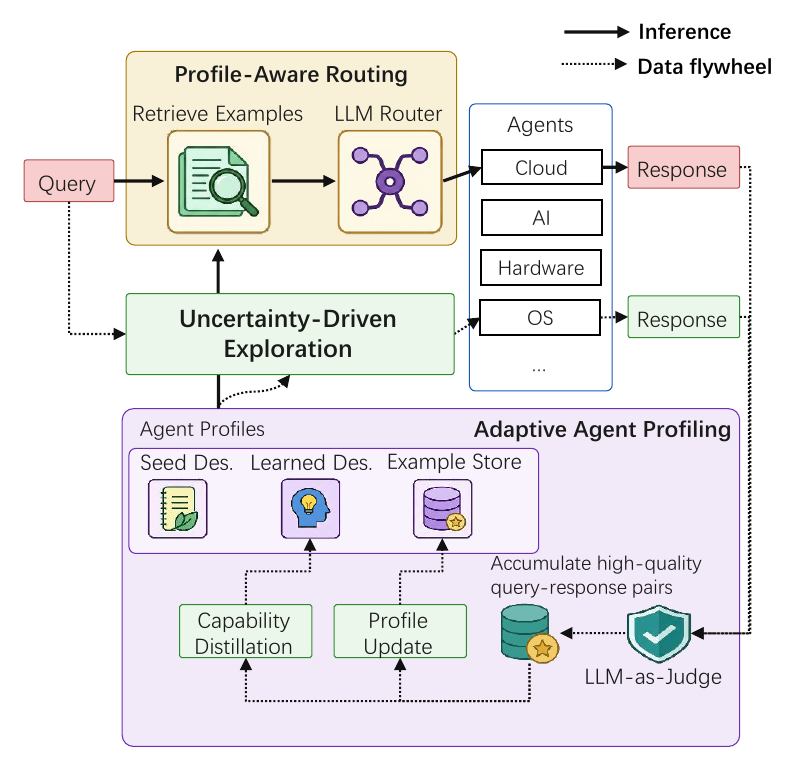}
  \caption{FlyRoute closed-loop data flywheel: adaptive profiling, uncertainty-driven exploration, and profile-aware routing.}
  \label{fig:pipeline}
\end{figure}

In this paper, we propose FlyRoute, a self-evolving agent-profiling framework that turns deployment interactions into continuously updated routing evidence. Our main contributions are:
\begin{itemize}\setlength\itemsep{0.05em}
    \item We identify and formalize the problem of continual agent profiling: constructing and continuously updating agent capability descriptions from deployment interactions, rather than relying on static developer-provided descriptions.
    \item We propose a profile-coverage-aware exploration policy that combines profile uncertainty, lexical query--agent relevance, and redundancy-aware novelty to efficiently collect evidence for continual profile refinement.
    \item Empirically, FlyRoute raises overall routing accuracy under the same backbone from 72.57\% (zero-shot LLM router) to 78.04\% at cold start ($k{=}5$ seeds per agent) and further to 89.83\% after streaming all 7{,}211 training queries through the flywheel, with the router consistently dispatching queries to the best-matching expert agent across four enterprise deployment domains.
\end{itemize}

\section{Related Work}
\label{sec:related}

\paragraph{Performance-Based Routing.}
The router primarily allocates requests across models to balance quality, cost, and latency. Prior work builds cascades and budget-aware selectors~\citep{chen2023frugalgpt,ding2024hybridllm}, benchmark- or preference-trained routing functions~\citep{shnitzer2023llmrouting,ong2024routellm}, compatibility embeddings~\citep{zhuang2025embedllm,chen2024routerdc}, universal or graph-based model-fleet routers~\citep{jitkrittum2025universal,feng2025graphrouter}, online or confidence-aware allocation policies~\citep{wang2025mixllm,chuang2025confidence,zhang2025uncertaintyrouting}, fusion-style systems that rank or blend candidate outputs~\citep{jiang2023llmblender,piskala2025dynamic}, and  probe-derived in-context vector representations for retraining-free model selection~\citep{wang2026iclrouter}. These methods primarily focus on allocating requests across a fixed pool of models to optimize quality, latency, or cost. While some support online adaptation of routing policies, they generally treat capability representations as fixed or externally maintained. FlyRoute instead studies continual profile enrichment, where capability evidence is accumulated directly from routed interactions and periodically incorporated into the routing context through profile updates.

\paragraph{Task-Based Routing.}
This mechanism selects the most suitable specialist agent or domain expert for each query. Existing systems use supervised domain classifiers or reward-distilled routers~\citep{simonds2024modem,lu2024routing}, planner-based tool and specialist invocation~\citep{shen2023hugginggpt,yao2023react,wu2023autogen,hong2024metagpt}, embedding-similarity matching between queries and agent profiles, and LLM- or graph-enhanced routers that exploit preferences, system structure, or explicit rationales~\citep{tran2025archrouter,zhang2025agentrouter,zhao2026tcar}. Despite architectural differences, these methods typically assume that routing evidence is specified before deployment or updated outside the routing loop. FlyRoute instead studies continual profile enrichment, where agent capability profiles are updated directly from routed interactions as evidence accumulates. This setting is particularly relevant when agent competence boundaries shift after deployment due to prompt, tool, or model changes. Consequently, we focus on same-backbone comparisons that isolate the value of evolving profiles. Unlike static-profile routers that assume complete descriptions and fixed taxonomies at initialization, FlyRoute begins from only a few seed examples and supports incremental profile growth under a streaming protocol.

\paragraph{Agent Profiling and Capability Discovery.}
Compared to the extensive work on routing architectures, agent profiling has received comparatively less attention, particularly regarding how capabilities should be represented and maintained over time. Most existing systems rely on static assumptions: developers provide descriptions, exemplars, or metadata at registration, which remain largely unchanged after deployment. This assumption becomes increasingly fragile in enterprise settings where prompts, tools, and foundation models evolve continuously. FlyRoute treats profiling as a continual process rather than a one-time initialization step. Conceptually, our exploration mechanism borrows the budget-aware intuition of active learning and bandit methods~\citep{settles2012active,lattimore2020bandit}, but differs in objective: rather than selecting examples to improve a fixed classifier, FlyRoute selectively probes agents whose capability profiles remain under-specified and incorporates accepted interactions into continuously updated routing profiles.

\section{Method}

\label{sec:method}

\subsection{Overview}
\label{sec:overview}

FlyRoute follows the closed loop in Figure~\ref{fig:pipeline}. When a new agent joins, it registers with minimal information---a name, an optional seed description, and a few seed examples---and the system initializes an empty profile. For each incoming query, the routing module decides which agents to dispatch the query to, balancing exploitation (selecting agents with strong profiles) and exploration (dispatching to agents with uncertain profiles to gather more evidence). After the selected agents produce responses, a quality gate decides which query--agent pairs enter the success store; our default implementation instantiates this gate with LLM-as-Judge on the training stream. Periodically, the system distills accumulated evidence into a refined capability description. This creates a self-reinforcing cycle: better profiles lead to better routing, which in turn generates higher-quality feedback, ultimately resulting in even better profiles.

\paragraph{Task formulation.}
Each benchmark example pairs a query $q$ with exactly one supervising specialist $a^\star \in \mathcal{A} = \{a_1, \ldots, a_N\}$, where $\mathcal{A}$ is the set of $N$ registered expert agents and each $a_i$ denotes a domain-specific agent, and routing accuracy on the held-out test set measures the fraction of queries for which the router correctly predicts $a^\star$.At training-replay time FlyRoute still merges exploitation candidates with exploration probes into a bounded dispatch set $\mathcal{S} \subseteq \mathcal{A}$ so several experts may answer; the quality gate decides which $(q,\text{agent})$ pairs merit success-store updates while the annotated gold route $a^\star$ remains inaccessible to the routing and profiling pipeline and is used exclusively for evaluation. 

\subsection{Adaptive Agent Profiling}
\label{sec:profiling}

The agent profile is the core artifact maintained by FlyRoute: rather than relying on a static developer-provided description, each profile evolves continuously as the system accumulates deployment evidence, and every routing decision is made directly from the current profile state.

\paragraph{Profile Structure.} Each agent $a_i$ maintains a profile $\mathcal{P}_i$ consisting of three elements:
\begin{itemize}\setlength\itemsep{0.05em}
    \item \textbf{Seed description} $d_i^{\text{seed}}$: the natural-language description provided by the developer at registration. This may be incomplete or inaccurate, but it serves as the initial signal for cold-start routing.
    \item \textbf{Learned description} $d_i^{\text{learn}}$: a refined capability description automatically distilled from accumulated success examples. This is initially empty and is generated once the agent has accumulated enough evidence.
    \item \textbf{Success example store} $\mathcal{E}_i = \{(q_j, r_j, s_j)\}$: a collection of query--response pairs with quality scores $s_j \geq \theta$, where $\theta$ is the quality threshold. Here, $q_j$ denotes the user query, $r_j$ the agent response, and $s_j$ the quality score assigned by the quality gate (LLM-as-Judge in our implementation).
    
\end{itemize}

\paragraph{Cold Start.} When a new agent $a_i$ registers, the system requires only a name, an optional seed description (which may be a rough one-sentence summary), and a small number of seed examples (as few as 5). The profile is initialized as $\mathcal{P}_i = (d_i^{\text{seed}}, \emptyset, \mathcal{E}_i^{\text{seed}})$, where $\mathcal{E}_i^{\text{seed}}$ contains the seed examples. This design intentionally minimizes onboarding requirements and reflects realistic deployments where only coarse descriptions and a handful of representative examples are available at registration time, and the router relies on the seed description until enough evidence is collected.

\paragraph{Profile Update.} After a query $q$ is dispatched to agent $a_i$ and a response $r_i$ is generated, a quality gate assigns a score $s_i \in [0, 1]$. If $s_i \geq \theta$, the triple $(q, r_i, s_i)$ is added to the success store:
\begin{equation}
    \mathcal{E}_i \leftarrow \mathcal{E}_i \cup \{(q, r_i, s_i)\}
\end{equation}

\paragraph{Capability Distillation.}
As the success store grows, raw examples become too numerous to use directly in routing prompts. We periodically distill the accumulated evidence into a refined capability description. Specifically, every $M$ new success examples, we prompt an LLM with the agent's seed description (optionally the prior learned summary), recent successes, and request an updated learned description $d_i^{\text{learn}}$ that captures what the agent actually excels at based on empirical evidence---possibly differing from the developer's registered description. The distillation prompt emphasizes summarizing evidence-supported capabilities rather than introducing unsupported ones. Distillation therefore serves as profile compression, producing a compact capability summary that complements retrieved success examples during routing.

\subsection{Profile-Aware Routing}
\label{sec:routing}

The routing module uses the adaptive profiles to make dispatch decisions. For each incoming query $q$, the router receives:
\begin{itemize}\setlength\itemsep{0.05em}
    \item The agent's current routing description $d_i = d_i^{\text{learn}}$ if available, otherwise $d_i^{\text{seed}}$, inserted verbatim (up to a length cap) into the LLM's system prompt as a per-agent profile summary.
    \item The top-$k$ most relevant success examples from $\mathcal{E}_i$, retrieved by BM25 lexical similarity to $q$, listed in the same prompt as query-local few-shot evidence.
\end{itemize}
The LLM is instructed to combine the global profile summaries---which approximate coverage over the whole success store after distillation---with the retrieved examples that focus on surface similarity to $q$. This differs from using retrieval alone: when BM25 returns weak or empty hits early in training, the profile summary still carries coarse-grained capability boundaries learned from the stream.

During the cold-start phase, the router relies on the seed description in the prompt together with sparse retrieved seeds. As the flywheel accumulates evidence, the distilled description replaces the seed in the active slot and sharpens the global signal seen by the LLM, while the example pool grows for BM25.

\subsection{Uncertainty-Driven Exploration}
\label{sec:exploration}

A naive approach to building agent profiles is to broadcast every query to all agents, but this is wasteful---many queries are irrelevant to certain agents, and the cost of invoking every agent for every query is prohibitive. Conversely, always routing to the best-known agent (pure exploitation) starves new or updated agents of the evidence they need to build profiles. FlyRoute introduces an uncertainty-driven exploration strategy that targets profile construction efficiently.

\paragraph{Profile Uncertainty.} We use a lightweight uncertainty surrogate based on profile coverage to determine how much exploration pressure an agent should receive. agent $a_i$ should receive using profile uncertainty computed from its success-store size:
\begin{equation}
    \label{eq:uncertainty}
    U(a_i) = \frac{1}{1 + |\mathcal{E}_i|^{\alpha}}
\end{equation}
where $|\mathcal{E}_i|$ is the number of quality-gated successes and $\alpha$ controls the decay rate. A profile backed by few examples has high $U(a_i)$, favoring exploration; as $|\mathcal{E}_i|$ grows, $U(a_i)$ decreases and the agent shifts toward exploitation in the composite scores below.

\paragraph{Exploration Value.} For a query $q$ and an agent $a_i$ not already selected by exploitation, we first define a base exploration value:
\begin{equation}
    \label{eq:explore}
    V_{\text{explore}}(q, a_i) = U(a_i) \cdot R(q, a_i)
\end{equation}
where $U(a_i)$ is the profile uncertainty from Eq.~\eqref{eq:uncertainty}, and $R(q, a_i)$ is the query--agent relevance estimated by lexical retrieval. Concretely, we index each agent's success-store queries with Okapi BM25~\citep{robertson2009bm25} and define $R(q, a_i)$ as the retriever's normalized relevance score for agent $a_i$ (higher when $q$ matches that agent's stored queries). This base score uses relevance as a gate, so uncertain agents are explored only when the incoming query is plausibly within their capability region. 

\paragraph{Novelty Reweighting.} The base score above filters out unrelated agents, but high lexical overlap can also indicate redundancy: dispatching another near-duplicate query may add little new profile evidence. We therefore reuse the normalized relevance signal from Eq.~\eqref{eq:explore} to derive a lightweight redundancy-aware weight:
\begin{equation}
    \label{eq:novelty}
    N(q, a_i) = 1 - R(q, a_i)
\end{equation}

The final exploration score is therefore a relevance-gated novelty reweighting:
\begin{equation}
    \label{eq:exploreplus}
    V_{\text{explore}}^{+}(q, a_i) = U(a_i) \cdot R(q, a_i) \cdot \bigl(1 + \beta \cdot N(q, a_i)\bigr)
\end{equation}
where $\beta$ controls how strongly we favor less redundant candidates among relevance-matched agents. 

\paragraph{Routing Decision.} The full routing procedure operates as follows:
\begin{enumerate}\setlength\itemsep{0.05em}
    \item \textbf{Exploitation}: Use the current router (an LLM with each agent's active profile description and top-$k$ retrieved success examples) to select a candidate agent set $\mathcal{S}_{\text{exploit}}$.
    \item \textbf{Exploration}: For each agent $a_i \notin \mathcal{S}_{\text{exploit}}$, compute $V_{\text{explore}}^{+}(q, a_i)$. Select up to $n_{\text{explore}}$ agents with the highest scores among those with $V_{\text{explore}}^{+} \ge \gamma$; if that set is empty, fall back to the top-$n_{\text{explore}}$ agents by $V_{\text{explore}}^{+}$ (so exploration does not stall when $\gamma$ is tight relative to the composite score scale).
    \item \textbf{Merge}: $\mathcal{S} = \mathcal{S}_{\text{exploit}} \cup \mathcal{S}_{\text{explore}}$.
\end{enumerate}
This strategy ensures that newly onboarded agents with uncertain profiles receive targeted query exposure, while irrelevant queries are not wasted on agents that cannot benefit from them.

\section{Experiments}
\label{sec:exp}

\subsection{Experimental Setup}

\paragraph{Dataset and Agents.} We evaluate FlyRoute on a proprietary enterprise benchmark collected from real routed user queries. The benchmark covers four business domains---Cloud Services, AI Accelerator, Server Hardware, and Mobile OS---with the training and test statistics shown in Table~\ref{tab:dataset}. We register one expert agent per domain; each agent starts with only a short developer-authored capability description and $k$ seed queries, so the initial profiles are deliberately sparse.

\begin{table}[t]
\caption{Dataset statistics for our enterprise developer-support benchmark.}
\label{tab:dataset}
\centering
\small
\begin{tabular}{lcc}
\toprule
Domain & Train & Test \\
\midrule
Cloud Services & 3,631 & 610 \\
AI Accelerator & 946 & 292 \\
Server Hardware & 1,044 & 294 \\
Mobile OS & 1,590 & 102 \\
\midrule
Total & 7,211 & 1,298 \\
\bottomrule
\end{tabular}
\end{table}

\paragraph{Baselines.}
We use same-backbone comparisons to isolate the value of evolving profiles. The LLM Router baseline uses Qwen3-8B to dispatch queries from static seed descriptions only, without retrieval or profile updates. FlyRoute (cold) uses the same BM25+LLM router as the full system before any training query is replayed, so each success store contains only the $k$ registration seeds.

\paragraph{Router configuration.}
Unless stated otherwise, all experiments use BM25 (top-20 retrieved queries per agent, truncated to 450 characters) in a few-shot routing prompt served by Qwen3-8B. Key hyperparameters are $\theta{=}0.7$, $\alpha{=}0.5$, $\beta{=}0.5$, $\gamma{=}0.06$, $n_{\text{explore}}{=}2$; full details and prompts are in Appendix~\ref{app:impl}--\ref{app:router_prompt}. The router backbone remains fixed throughout all experiments.

\paragraph{Streaming Protocol.}
We simulate cold start with $k{=}5$ seed queries per agent and replay the remaining training data as a stream. For each query, FlyRoute routes using the current profiles, adds exploration targets via Eq.~\eqref{eq:exploreplus}, and writes quality-gated interactions into the corresponding success stores. The acceptance gate is implemented with LLM-as-Judge (Appendix~\ref{app:router_prompt}). Gold routing annotations are hidden from the router, judge, and profile-construction pipeline and are used only for evaluation. All checkpoints are evaluated on the held-out test set under identical settings.

\paragraph{Metrics.} Training and test splits provide exactly one gold expert per query. Overall accuracy is the fraction of test queries whose dispatch matches the supervising gold route. Per-domain columns report the accuracy restricted to queries whose gold annotation names that domain. 

\subsection{Main Results}

Table~\ref{tab:main} shows that FlyRoute improves routing accuracy. Seed-only retrieval already outperforms the zero-shot LLM Router (78.04\% vs.\ 72.57\%), while the full flywheel reaches 89.83\% after replaying the training stream (+17.26pp over zero-shot, +11.79pp over cold start).

\begin{table}[t]
\begin{threeparttable}
\caption{Main results on our enterprise developer-support dataset. $^\dagger$: BM25 + LLM-fewshot (Qwen3-8B). Cold: seed-only stores before streaming. Flywheel: after all training queries.}
\label{tab:main}
\centering
\small
\begin{tabularx}{\columnwidth}{LCCCCC}
\toprule
Method & Overall & CS & AIA & SHW & MOS \\
\midrule
LLM Router & 72.57 & 93.61 & 62.33 & 37.41 & 77.45 \\
\midrule
FlyRoute (cold)$^\dagger$ & 78.04 & 82.30 & 78.42 & 69.05 & 77.45 \\
\midrule
FlyRoute (flywheel)$^\dagger$ & \textbf{89.83} & \textbf{94.26} & \textbf{88.70} & \textbf{81.29} & \textbf{91.18} \\
\bottomrule
\end{tabularx}
\begin{tablenotes}[flushleft]
\footnotesize
\item[] \emph{Domain abbreviations:} \textbf{CS}, Cloud Services; \textbf{AIA}, AI Accelerator; \textbf{SHW}, Server Hardware; \textbf{MOS}, Mobile OS.
\end{tablenotes}
\end{threeparttable}
\end{table}

The comparison separates retrieval-augmented cold start from self-evolution: seed examples provide immediate lexical grounding, while the flywheel expands $\mathcal{E}_i$, probes underspecified agents, and refreshes profile summaries as more traffic is observed.

\subsection{Data Flywheel Effect}

To measure whether routing improves as profile evidence accumulates, we track held-out accuracy after increasing amounts of streamed training data. Table~\ref{tab:flywheel} reports checkpoints from the initial five-seed state through the full 7,211-query stream.

\begin{table}[t]
\begin{threeparttable}
\caption{Data flywheel effect: held-out routing accuracy (\%) after streaming $N$ training queries. Row $\Delta$: gain from $N{=}0$ to $N{=}7{,}211$.}
\label{tab:flywheel}
\centering
\small
\begin{tabularx}{\columnwidth}{rCCCCC}
\toprule
Queries & Overall & CS & AIA & SHW & MOS \\
\midrule
0 & 78.04 & 82.30 & 78.42 & 69.05 & 77.45 \\
500 & 84.05 & 93.61 & 83.56 & 64.63 & 84.31 \\
1{,}000 & 85.13 & 92.79 & 85.62 & 68.71 & 85.29 \\
2{,}000 & 85.75 & 91.80 & 86.64 & 72.11 & 86.27 \\
4{,}000 & 88.44 & 93.77 & 86.99 & 78.57 & 89.22 \\
7{,}211 & \textbf{89.83} & \textbf{94.26} & \textbf{88.70} & \textbf{81.29} & \textbf{91.18} \\
\midrule
$\Delta$ & +11.79 & +11.96 & +10.28 & +12.24 & +13.73 \\
\bottomrule
\end{tabularx}
\begin{tablenotes}[flushleft]
\footnotesize
\item[] \emph{Abbreviations:} Table~\ref{tab:main}.
\end{tablenotes}
\end{threeparttable}
\end{table}

Routing accuracy improves steadily as profile evidence accumulates, increasing from 78.04\% at cold start to 89.83\% after the full replay stream (+11.79pp). All domains improve by the final checkpoint.

\subsection{Exploration Strategy Comparison}

We next isolate the exploration policy by holding the router, seed profiles, streaming order, and LLM-as-Judge acceptance gate fixed. Table~\ref{tab:exploration} reports held-out accuracy after 500 streamed training queries, when profiles are still sparse enough for exploration quality to matter.

\begin{table}[t]
\caption{Exploration strategy comparison after 500 streamed training queries. $\Delta$: held-out overall accuracy gain from $N{=}100$ to $N{=}500$.}
\label{tab:exploration}
\centering
\small
\begin{tabular}{lcc}
\toprule
Strategy & Final Acc.\ (\%) & $\Delta$ (pp) \\
\midrule
No Exploration & 81.05 & +3.01 \\
$\epsilon$-Greedy ($\epsilon{=}0.3$) & 81.90 & +3.86  \\
Random Broadcasting & 81.97 & +3.93 \\
\textbf{FlyRoute ($k$=5, flywheel)} & \textbf{84.05} & \textbf{+6.01} \\
\bottomrule
\end{tabular}
\end{table}

At $N{=}500$, FlyRoute reaches 84.05\%, outperforming random broadcast (81.97\%), $\epsilon$-greedy exploration (81.90\%), and pure exploitation (81.05\%). It suggests that our strategy is more effective, better than both random broadcasting and $\epsilon$\nobreakdash-greedy. Late-stream results are discussed in Appendix~\ref{app:impl}.

\subsection{Ablation Study}

Table~\ref{tab:ablation} ablates the main sources of routing evidence after the full training stream, keeping the seed setting, backbone router, and metric fixed.

\begin{table}[t]
\begin{threeparttable}
\caption{Ablation study after the full training stream.}
\label{tab:ablation}
\centering
\small
\begingroup
\setlength{\tabcolsep}{2.5pt}
\begin{tabularx}{\columnwidth}{@{}>{\raggedright\arraybackslash}p{0.2\columnwidth}*{5}{>{\centering\arraybackslash}X}@{}}
\toprule
Config. & Overall & CS & AIA & SHW & MOS \\
\midrule
FlyRoute (flywheel) & \textbf{89.83} & \textbf{94.26} & \textbf{88.70} & \textbf{81.29} & \textbf{91.18} \\
w/o Novelty & 89.06 & 94.43 & 88.36 & 79.59 & 86.27 \\
w/o Distill. & 89.29 & 94.43 & 86.99 & 81.63 & 87.25 \\
w/o Explore & 88.52 & 93.44 & 88.70 & 77.89 & 89.22 \\
w/o Judge & 88.44 & 91.15 & 91.44 & 79.93 & 88.24 \\
\bottomrule
\end{tabularx}
\endgroup
\begin{tablenotes}[flushleft]
\footnotesize
\item[] \emph{Abbreviations:} Table~\ref{tab:main}.
\end{tablenotes}
\end{threeparttable}
\end{table}

The largest degradations arise from removing the quality gate (1.39pp) or exploration (1.31pp), indicating the importance of reliable evidence collection and profile coverage. Removing novelty reweighting or capability distillation yields smaller but consistent drops, suggesting that both redundancy control and profile summarization provide complementary benefits beyond retrieval alone.

\section{Conclusion}
\label{sec:conclusion}

We presented FlyRoute, a continual agent-profiling framework that replaces static capability descriptions with profiles refined from routed interactions. FlyRoute combines quality-gated evidence collection, capability distillation, retrieval-augmented profiling, and targeted exploration while keeping the routing backbone fixed. On our benchmark, FlyRoute improves routing accuracy from 72.57\% to 78.04\% with five seed examples and further to 89.83\% after replaying 7,211 training queries. These results suggest that maintaining agent profiles as evolving artifacts can substantially improve routing quality without retraining the router.

\section*{Limitations}
Our evaluation is conducted on proprietary enterprise developer-support logs spanning four domain-specific expert routes. While this setting reflects a realistic industrial deployment, it limits external reproducibility and does not establish transferability to other industries, languages, open-domain assistants, or substantially larger agent ecosystems. Future work should evaluate FlyRoute on publicly available routing benchmarks and more diverse agent populations.

A second limitation concerns the distinction between profile evolution and agent evolution. FlyRoute is motivated by real deployments in which agent capabilities change over time through prompt updates, tool additions, or model replacements. However, our experiments evaluate continual profile refinement under a largely stationary set of expert agents rather than performing controlled capability-shift interventions. Demonstrating adaptation under explicit prompt, tool, or model changes remains an important direction for future work.

Finally, FlyRoute relies on the quality of its feedback signal. In our default setup, profile updates are driven by an LLM-as-Judge quality gate. Although accepted interactions are filtered through a threshold and aggregated over time, routing performance may still be sensitive to judge calibration, prompt design, and model choice. Further study is needed to understand robustness under alternative judging strategies and noisy feedback conditions.

\bibliography{custom}

\appendix
\section{Illustrative Training Queries}
\label{app:data_samples}

\noindent Table~\ref{tab:app-train-samples} illustrates typical user questions in our enterprise developer-support benchmark (\S\ref{sec:exp}). We include one example per gold domain, drawn at random within that domain from the training split so coverage is balanced across the four experts. Each row reflects the single-label supervision used in our experiments. Wording is \textbf{paraphrased in English} and lightly redacted for publication; it may differ from raw logs.

\begin{table*}[t]
\centering
\small
\begin{tabular}{@{}p{0.62\textwidth}p{0.32\textwidth}@{}}
\toprule
\textbf{Query (English paraphrase)} & \textbf{Gold domain} \\
\midrule
How should a Gearbox workflow elastically scale a Deadline render cluster when driving Maya batch rendering on the cloud? & Cloud Services \\
How do I obtain the base OS image (environment setup for device toolchain)? & AI Accelerator \\
What Apache Traffic Server (ATS) version is recommended on Kunpeng-related server stacks, and why? & Server Hardware \\
What are compatibility requirements and usage steps for cross-device clipboard features on HarmonyOS-class client OS? & Mobile OS \\
\bottomrule
\end{tabular}
\caption{Illustrative training-split queries (one per domain). English paraphrases for readability.}
\label{tab:app-train-samples}
\end{table*}

\section{Implementation Details}
\label{app:impl}

The router is a BM25 retriever over each agent's stored success queries coupled with Qwen3-8B (API access) under a few-shot prompt design. We retrieve the top 20 queries per agent for scoring and prompt construction. The active profile text ($d_i^{\text{learn}}$ if distilled, otherwise $d_i^{\text{seed}}$) is placed in the system prompt together with retrieved snippets, each truncated to 450 characters. Capability distillation calls the same model after every $M{=}20$ new accepts. Exploration uses $\alpha{=}0.5$, quality threshold $\theta{=}0.7$, $\beta{=}0.5$, and $\gamma{=}0.06$ applied to $V_{\text{explore}}^{+}$. We add at most $n_{\text{explore}}{=}2$ exploration agents per query, with rank-based fallback when too few pass $\gamma$, and treat a profile as well explored once $|\mathcal{E}_i|\!\geq\!10$. Unless otherwise noted, we stream the full training split in chronological order. Training-time acceptance uses LLM-as-Judge scores from the same Qwen3-8B backbone ($s_i \geq \theta$); optional diagnostics replace that judge with deterministic acceptance from the annotated route, eliminating judge variance whenever labels exist. Capability distillation uses the user-message template in Figure~\ref{fig:distill_prompt}. The LLM-as-Judge gate uses Figure~\ref{fig:judge_prompt}.

\paragraph{Additional experiment notes.} The exploration-strategy comparison in Table~\ref{tab:exploration} stops at 500 training queries because early profile construction is where exploration policies are most distinguishable. Continuing the same sweep through all 7,211 queries narrows the gaps among substitutes: no exploration and $\epsilon$-greedy both finish at 88.52\% routing accuracy, while random broadcast reaches 88.06\%. This convergence is expected because exploitation and large retrieval stores dominate once many accepted examples have accumulated. In the distillation ablation, the modest aggregate drop should not be read as evidence that distillation is unnecessary: under strong BM25 retrieval and large success stores, retrieved lexical neighbors already carry much of the signal, while distilled summaries remain useful for compact routing context and domains with weaker seed coverage.

\section{Prompt Design}
\label{app:router_prompt}

To demonstrate the exact instructions used in our system, we present the prompts that guided routing, distilled profile updates, and training-time acceptance. Figures~\ref{fig:system_prompt}--\ref{fig:judge_prompt} summarize the routing \texttt{system} prompt, the capability-distillation prompt, and the LLM-as-Judge prompts aligned with {\small\texttt{dataset/llm\_as\_judge.py}} (English translations of the deployed Chinese templates).

\begin{figure*}[!t]
\centering
\begin{tcolorbox}[
  enhanced,
  width=\linewidth,
  colback=myblue!5!white,
  colframe=myblue!75!black,
  boxrule=0.55pt,
  arc=2.2mm,
  titlerule=0pt,
  toptitle=1.2mm,
  bottomtitle=1.2mm,
  left=10pt,
  right=10pt,
  top=9pt,
  bottom=9pt,
  lefttitle=4mm,
  fonttitle=\bfseries\small,
  title={FlyRoute router --- \texttt{system} prompt (\textit{English translation})}
]
\small\setlength{\parindent}{0pt}\setlength{\parskip}{0.4em}
\medskip
You are an expert at routing queries. Combine the \emph{Agent profile summary} with the \emph{BM25 retrieval examples}:

The profile gives a global summary of successful trajectories; the retrieval lines highlight past successes lexically close to the current user query.
\vspace{10pt}

\textbf{[Initial developer descriptions]}

1. Cloud Services: Cloud product resources, tools, services, specifications, purchase, documentation, etc.

2. AI Accelerator: Ascend stack, CANN, MindSpore, operators, model training and inference tooling, etc.

3. Server Hardware: Kunpeng, openEuler, DevKit, BoostKit, HPC, acceleration libraries, storage, etc.

4. Mobile OS: HarmonyOS, Ability Kit, ArkTS, ArkUI, HarmonyOS~Next, etc.

\vspace{10pt}
\textbf{[Agent profile summary]}

Below are global capability statements for each route (seed text at registration; after distillation, summaries induced from historical successes, to capture shared competence beyond the retrieved examples):

\nopagebreak\texttt{\{profile\_block\}}

\vspace{10pt}
\textbf{[BM25 retrieval examples]} (successful queries close to the current query and their supervising route label)

\texttt{\{examples\_section\}}

\vspace{10pt}
\textbf{[Output rules]}

1. Output exactly one agent dispatch per query: choose one of Cloud Services, AI Accelerator, Server Hardware, Mobile OS defined in the deployment template.

2. Output route names only---no explanation or extra text.
\end{tcolorbox}
\caption{Prompt for FlyRoute. Notice that the live system prompt uses Chinese-language tokens; English labels shown here for readability.}
\label{fig:system_prompt}
\end{figure*}

\begin{figure*}[!t]
\centering
\begin{tcolorbox}[
  enhanced,
  width=\linewidth,
  colback=myblue!5!white,
  colframe=myblue!75!black,
  boxrule=0.55pt,
  arc=2.2mm,
  titlerule=0pt,
  toptitle=1.2mm,
  bottomtitle=1.2mm,
  left=10pt,
  right=10pt,
  top=9pt,
  bottom=9pt,
  lefttitle=4mm,
  fonttitle=\bfseries\small,
  title={FlyRoute capability distillation --- \texttt{user} prompt (\textit{English translation})}
]
\small\setlength{\parindent}{0pt}\setlength{\parskip}{0.4em}
You are an expert at analyzing AI agent capabilities. From the information below, produce a \textbf{concise} capability description for the agent.

\vspace{10pt}
\textbf{[Agent id]}

\texttt{\{agent\_id\}}

\vspace{10pt}
\textbf{[Developer's initial description]}

\texttt{\{seed\_desc\}} \quad (\textit{if missing, code fills a fixed empty-valued sentinel string})

\texttt{\{prev\_learned\_block\}}

\vspace{10pt}
\textbf{[Recent successful user queries handled by this agent (excerpt, at most 30)]}

\texttt{\{examples\}} \quad (\textit{numbered lines; code inserts a placeholder line if the list is empty})

\vspace{10pt}
\textbf{[Requirements]}

1. Ground the summary only in queries that actually succeeded; do not fabricate capabilities.

2. If empirical behavior differs from the initial description or the prior distilled summary, follow the newest evidence; retain prior wording that still holds and revise only when evidence conflicts.

3.Keep the description concise: $\leq$500 characters in the deployed Chinese template (We state these rules in English for readability).

4. If no prior distilled description exists, \texttt{\{prev\_learned\_block\}} is omitted.

\vspace{10pt}
\textbf{[Output format]}

\texttt{<Description>}\textit{\ldots capability text (within the character cap) \ldots}\texttt{</Description>}
\end{tcolorbox}
\caption{Distillation prompt for FlyRoute. Notice that the live template is in Chinese; we translate it to English here for readability.}
\label{fig:distill_prompt}
\end{figure*}

\begin{figure*}[!t]
\centering
\begin{tcolorbox}[
  enhanced,
  width=\linewidth,
  colback=myblue!5!white,
  colframe=myblue!75!black,
  boxrule=0.55pt,
  arc=2.2mm,
  titlerule=0pt,
  toptitle=1.2mm,
  bottomtitle=1.2mm,
  left=10pt,
  right=10pt,
  top=9pt,
  bottom=9pt,
  lefttitle=4mm,
  fonttitle=\bfseries\small,
  title={FlyRoute LLM-as-Judge --- prompts (\textit{English translation})}
]
\small\setlength{\parindent}{0pt}\setlength{\parskip}{0.4em}

\textbf{system prompt}

You are an expert in intent understanding. You evaluate response quality to determine whether a query--response pair should be accepted into a success-store exemplar corpus.

You receive a paired input: the user's factual query $q$ and candidate response $r$ from a routed expert. $r$ may be partial, vague, produced by tooling or LLMs (including placeholders, refusals, or missing answers). Your task is not to rewrite the answer but to assign a scalar score expressing whether accepting $(q, r)$ into a ``success-store'' exemplar corpus is warranted.

\vspace{6pt}
\textbf{[Business-scope boundaries]} (consistent with four expert routes)\\
1. Cloud Services: Cloud product resources, tools, services, specifications, purchase, documentation, etc.\\
2. AI Accelerator: Ascend stack, CANN, MindSpore, operators, model training and inference tooling, etc.\\
3. Server Hardware: Kunpeng, openEuler, DevKit, BoostKit, HPC, acceleration libraries, storage, etc.\\
4. Mobile OS: HarmonyOS, Ability Kit, ArkTS, ArkUI, HarmonyOS Next, etc.

\vspace{6pt}
\textbf{[Scoring principles]}\\
1. $r$ must address the technical subject of $q$; off-topic chatter, unrelated marketing, or empty acknowledgments lowers the score.\\
2. If $r$ sprawls largely outside those four competencies with weak grounding, penalize under ``off-topic/low applicability.''\\
3. Responses that are blank, boilerplate refusals, or contain no substantive guidance should score $\leq 0.35$.\\
4. Partially helpful but misses the crux earns mid-range scores; only concrete, actionable, mostly non-contradictory answers earn high scores (watch for hallucination).\\
5. The FlyRoute gate typically applies $\theta \approx 0.70$: slightly beyond ``barely usable'' denotes evidence worth caching.

\vspace{6pt}
\textbf{[Optional calibration block]} Few-shot excerpts can be inlined here via \texttt{\{examples\_optional\}} when enabled in code.

\vspace{6pt}
\textbf{[Output schema]} Return a single JSON object. Use ASCII quotes only; omit Markdown wrappers. The mandatory key is \texttt{quality\_score}, whose value must be a float in $[0.0, 1.0]$ (two decimal places optional). Optionally add \texttt{reason}; parsers consume only \texttt{quality\_score}.

\vspace{6pt}
\textbf{user message template}

(When available) This response is attributed to route \texttt{\{routed\_agent\_id\}} (deployment tokens: \texttt{Cloud-Services}, \texttt{AI-Accelerator}, \texttt{Server-Hardware}, \texttt{Mobile-OS}). Consider whether $r$ reads like domain-plausible developer guidance for that route.

\vspace{6pt}
\textbf{[Query]}

\texttt{\{query\}}

\vspace{6pt}
\textbf{[Candidate response]}

\texttt{\{response\}}

\vspace{6pt}
Output JSON obeying the system instructions.
\end{tcolorbox}
\caption{LLM-as-Judge prompts for FlyRoute. The deployed templates are Chinese; we convert them to English for readability.}
\label{fig:judge_prompt}
\end{figure*}

\begin{figure*}[!t]
\centering
\begin{tcolorbox}[
  enhanced,
  width=\linewidth,
  colback=myblue!5!white,
  colframe=myblue!75!black,
  boxrule=0.55pt,
  arc=2.2mm,
  titlerule=0pt,
  toptitle=1.2mm,
  bottomtitle=1.2mm,
  left=10pt,
  right=10pt,
  top=9pt,
  bottom=9pt,
  lefttitle=4mm,
  fonttitle=\bfseries\small,
  title={FlyRoute Learned descriptions (\textit{English translation})}
]
\small\setlength{\parindent}{0pt}\setlength{\parskip}{0.4em}

\vspace{10pt}
\textbf{[Cloud Services]}

The agent provides broad cloud resource management and service coverage across compute, storage, networking, databases, security, and DevOps. It supports configuration, monitoring, and management of resources such as RDS, Redis, OBS, and GaussDB (DWS), including usage/cost reporting, SKU queries, automated deployment, and resource reclamation. It supports IAM, service provisioning, API guidance, and SDK operations, and can assist with database migration, certificate management, container configuration, and code checks. It supports services such as CCE, CDN, CodeArts Check, and container image services—from resource creation through permission assignment. It answers questions on cloud product capabilities, deployment, troubleshooting, and API usage for developers, operators, and enterprise workloads. It supports OCR, FRS, SparkRTC, and related services, and provides cost breakdowns, capacity planning, security auditing, and automated operations for enterprise cloud management.

\vspace{10pt}
\textbf{[AI Accelerator]}

The agent answers questions on the Ascend AI platform: CANN, MindSpore, operator development, and training/inference toolchains. It supports deploying and configuring PyTorch, MindSpore, and similar frameworks on bare metal and in Docker, with guidance on environment setup, device attach, and network parameters. It supports Ascend Deployer, MindCluster, MindEdge, and related components—logging, resource queries, and mode switching. It supports virtualized instances and Atlas inference configuration and tuning (e.g., vNPU management, VDEC core counts). It documents interfaces such as dcmi\_subscribe\_fault\_event and alarm resource queries, including behavior, parameters, and constraints. It supports common troubleshooting such as missing modules and memory-interface adjustments, with strong toolchain and systems integration. It clarifies API behavior, parameters, and constraints—including PyTorch native API coverage and precision\_tool deployment—and supports plugin development, SDK configuration, version compatibility checks, and error-code retrieval for training, inference, deployment, and operations.

\vspace{10pt}
\textbf{[Server Hardware]}

The agent focuses on Kunpeng processors, the openEuler OS, the DevKit toolchain, the BoostKit acceleration suite, and HPC. It provides end-to-end support for development and debugging, system deployment, performance optimization, and security configuration. It supports scenarios such as Kunpeng–x86 difference analysis, acceleration-library integration, network and I/O tuning, compiler optimization, MPI communication tuning, GPU performance tuning, memory management, and NUMA-node analysis. It can handle technical issues including header parsing, tool usage, compile verification, instruction substitution, platform difference analysis, file permissions, authentication service configuration, and storage-system debugging. It has successfully addressed concrete problems such as TEEC API usage, Hyper MPI configuration, KVSIP complex arithmetic, compiler option tuning, Kbox container information collection, Ceph RPM builds, GPU performance optimization, Jenkins cluster setup, TrustZone RSA operations, BoostKit feature configuration, OpenStack troubleshooting, and secure-OS testing. It spans the full chain of development, operations, and performance tuning, and is suited to technical consulting and issue resolution in the Kunpeng ecosystem.

\vspace{10pt}
\textbf{[Mobile OS]}

The agent focuses on full-stack HarmonyOS development: system architecture, API usage, component development, and platform rules. It supports struct parsing and functional guidance for ArkTS, ArkUI, Ability Kit, ArkGraphics, and related modules; API change lookup; migration guidance; and HarmonyOS NEXT adaptation plans. It supports device adaptation cases such as round-screen wearables and physical-keyboard input anomalies. It supports release configuration, SDK review rule interpretation, personal-data collection compliance, and marketplace qualification review. It can interpret structs such as OH\_RecorderInfo, ColorXY, and Rdb\_Statistic, and supports scenarios including Vulkan surface creation, audio encoder instantiation, two-way binding for Slider, and reuse patterns for ListItemGroup with LazyForEach. It applies to HarmonyOS 5.0.0 and NEXT, including Netstack certificate-chain verification, Connectivity Kit Bluetooth scanning, and ArkWeb navigation. It has handled diverse requests such as sensor access, payment merchant configuration, Image module queries, Crypto key derivation, AbilityDelegatorArgs usage, ArkTS array creation, Cangjie multithreading, API maintenance status, struct member parsing, in-app charging configuration, and error-code diagnosis.

\end{tcolorbox}
\caption{FlyRoute learned descriptions. The deployed templates are Chinese; we convert them to English for readability.}
\label{fig:learned_prompt}
\end{figure*}

\end{document}